\documentclass[runningheads]{llncs} 
\usepackage{graphicx}
\usepackage[utf8x]{inputenc}
\usepackage[letterpaper, margin=1in]{geometry}

\usepackage{support-caption}
\usepackage{subcaption}

\inputencoding{latin2}

\inputencoding{utf8}

\usepackage{tabularx}
\usepackage{makecell}

\begin{document}
%
\title{A framework for optimizing COVID-19 testing policy using a Multi Armed Bandit approach}
\titlerunning{A framework for optimizing COVID-19 testing policy}
%
\author{Hagit Grushka-Cohen \inst{1} \and Raphael Cohen \inst{2} \and Bracha Shapira \inst{1} \and Jacob Moran-Gilad \inst{3} \and Lior Rokach \inst{1}}
\authorrunning{H. Grushka-Cohen et al.}

\institute{Department of Software and Information Systems Engineering, Ben-Gurion University of the Negev,  Israel \and Chorus.ai, San Francisco, CA \and Department of Health Systems Management, School of Public Health, Faculty of Health Sciences, Ben-Gurion University of the Negev,  Israel}



\maketitle   
\begin{abstract}
Testing is an important part of tackling the COVID-19 pandemic. Availability of testing is a bottleneck due to constrained resources and effective prioritization of individuals is necessary. Here, we discuss the impact of different prioritization policies on COVID-19 patient discovery and the ability of governments and health organizations to use the results for effective decision making. We suggest a framework for testing that balances maximal discovery of positive individuals with the need for population-based surveillance aimed at understanding disease spread and characteristics. This framework draws from similar approaches to prioritization in the domain of cyber-security based on ranking individuals using a risk score and then reserving a portion of the capacity for random sampling. This approach is an application of Multi-Armed-Bandits maximizing exploration/exploitation of the underlying distribution. We find that individuals can be ranked for effective testing using a few simple features, and that ranking them using such models we can capture 65\% (CI: 64.7\%-68.3\%) of the positive individuals using less than 20\% of the testing capacity or 92.1\% (CI: 91.1\%-93.2\%) of positives individuals using 70\% of the capacity,  allowing reserving a significant portion of the tests for population studies. Our approach allows experts and decision makers to tailor the resulting policies as needed allowing transparency into the ranking policy and the ability to understand the disease spread in the population and react quickly and in an informed manner.

\end{abstract}
\section{Introduction}

The SARS-CoV-2 virus causing COVID-19 has spread globally and was declared as a pandemic by the World Health Organization. Testing for COVID-19 has been an important tool in both monitoring the spread of the virus as well as in the attempts of preventing its rapid spread. Quickly identifying outbreaks is a key to combating the spread of the disease \cite{cohen2020countries,salathe2020covid}. However testing is bounded by the available resources such as number of available tests, reagent bottlenecks, laboratory accreditation and technical constrains.
Since the number of tests is limited, governments and decision makers around the world apply policies guiding who should be tested. During the outbreak countries implemented different testing policies. These policies vary in a wide range of approaches such as: testing of symptomatic individuals only, testing of symptomatic individuals and their contacts, test only hospitalized patients \cite{salathe2020covid,ourworldindata2020testing}, test severe cases \cite{medicalxpress2020sweden}, test with a rigid policy for optimizing to the most likely phenotypes \cite{shoer2020should}, or testing with a policy and two random groups as done in Iceland \cite{gudbjartsson2020spread}. In the case of unlimited test capacity a policy of testing everyone was applied in Wuhan (Wuhan May-2020, 11M tests) \cite{wuhan2020may,wuhan202011million}. 

Testing serves multiple purposes. The first is to identify sick individuals in order to treat them correctly and safely. The second purpose is to stop the disease from spreading by contact tracing of positive individuals, this is critical for lowering the reproduction number of the disease. Finally, early identification of disease activity and case clusters allows governments to respond by increasing contact tracing, testing, and shelter in place or social distancing.

Most countries implement risk driven policies  \cite{shoer2020should} and result in data which is biased towards the patients that conform to the policy, as others, such as asymptomatic carries [find cite], will not get tested. This bias causes two major problems.  Firstly, decision makers do not have the information necessary to make decision about the virus spread (or lack accurate information). Lack of information prevent governments from properly assessing risks in the first stages of a new pandemic when virus characteristics, such as the death rate, were unknown and estimated based on biased information \cite{wu2020estimating}. Secondly, time to identify new disease clusters is prolonged, since most testing resources are spent on the existing clusters as symptomatic cases are prioritized ("searching under the lamppost"). 

In recommendation systems, this is a well known phenomenon, named "filter bubble", in which the users of the system are trapped in a subspace of options thereby losing the ability to explore beyond what they already know \cite{nguyen2014exploring,chen2019serendipity}. 

Monitoring a population with a limited capacity for testing individuals is not unique for epidemiology, a similar challenge exist in the realm of cyber-security where users are monitored for signs of compromise such as malware installation or data theft by employees. Database Activity Monitoring systems log and test user actions to identify compromised users and identify their actions. The purpose of data security is similar to the task of testing and contact tracing organizations. The sheer amount of daily user activity in IT database systems prevents testing and logging every action, instead these systems use an approach of ranking actions by risk using a policy and logging only potentially risky activity. This approach suffered from the same filter bubble problem of data biased towards the policy, and inability to quickly react to new threats or changes in activity.

In this paper we present a framework for data driven testing policy that optimizes both test utilization (i.e. maximizing detection of new cases) while investing resources in exploration that allows gathering unbiased surveillance information about the population and quickly identifying new disease clusters. This framework borrows from a similar approach based on Multi Armed Bandits (MAB) for balancing exploration and exploitation when monitoring database users.

To be adopted, approaches that enhance policies created by experts have to maintain the experts in the loop. The models must be explainable, no 'black box' decision can be made, especially with clinical outcomes and public health effects of such a system. The model also needs to be adjustable and the experts need a measure of control for tuning it, allow the experts to make decisions balancing different risks, and enable the experts to add features (such as adding geographic, clinical or population demographics features).

The system we propose allows the experts to make decisions that drive the ranking model, add features on the fly if needed and balance exploration/exploitation as needed. Using a data set of COVID-19 patient symptoms released by the Israeli Ministry of Health we find that it's possible to rank patients effectively for optimizing recall of testing, correlating to the exploitation mechanism of MAB. Using 10\% the testing capacity is enough in order to identify 66.6\%  of the positive individuals (CI: 64.7\%-68.3\%), and at 70\% of the capacity 92.1\%  (CI: 91.1\%-93.2\%) are identified. This enables diverting testing capacity towards exploration without harming recall and present a novel framework for COVID-19 testing policy. We trained two models on different time frames and find significant difference in their efficacy, suggesting policy changes affect what's learned from the data which provides another strong incentive for using a major part of the testing capacity for exploration such as random sampling.

\section{Background}

To maximize the detected infected individuals \cite{kaplan2020logistics}  suggest a screening approach that is made of combining three complementary testing strategies: (1) test all individuals with typical COVID-19 symptoms (2) deploy targeted testing based on the population density (to maximize the likelihood of infections) (3) reserve a small number of tests for random population screening to monitor the testing policy. They suggest combining it with campaign and funding for isolation of infected individuals. 
Iceland testing policy was also made of both :  targeted testings of individuals at high risk and population screening \cite{gudbjartsson2020spread}. Initially they invested in testing all individuals who desired to get tested, mostly from the capital metropolis. To provide a control testing group they randomly selected individuals who where summoned to be tested representing the age and gender demographics.\cite{shoer2020should} modeled the probability of individuals to test positive to SARS-CoV-2 based on their response to a daily symptoms survey. They did not suggest or evaluate the effect of using their predictive model for ranking.

Both policies lack the motive of dynamic updating of the policy. The first \cite{kaplan2020logistics} paper does suggest an estimation of infection prevalence however it can be viewed as special case of our suggested model where population density or epicenter is of high priority in the risk score model. The second \cite{gudbjartsson2020spread} work has both exploration and exploitation motives however the policy of who should be tested is static and there is no element of learning from one time frame to the following. We suggest a testing framework that enables maintaining and constantly updating the testing policy in a principled manner while keeping the experts and decision makers in the loop. This framework also enables introducing new features to the model. 

The Icelandic research \cite{gudbjartsson2020spread} also shows that targeted testing improves the tests' recall.  13.3\% of the targeted tests resulted  positive versus only 0.8\% from the "open invitation" population, and only 0.6\% recall for the randomly selected "control group",  which is compatible both with \cite{shoer2020should} and our results that shows a "risk score" can be learned and improve the test recalls.

\subsection{Assigning a Risk Score}

Many domains, such as health, insurance and bank loans use a risk score as a common practice for prioritizing. \cite{lloyd2004framingham,nelson1981apgar}. 
The risk score is not a policy, but can be used as part of the policy for decision making \cite{thomas2000survey}, e.g. newborns with low Apgar score will stay an extra day at the hospital\cite{jepson1991apgar}.
Risk score can be assigned based on predefined list of rules and conditions. The science community has been extensively working on using machine learning models for risk scoring in many domains \cite{ranganath2016deep,grushka2016cyberrank,evina2019enforcing} 
For an ML risk score to be accepted by the medical community it is important that it allows experts to tweak it (add clinical features, local biases) and is explainable \cite{lundberg2018explainable}. 
Clalit Health Services (CHS) created a risk-scoring tool to predict the severity of COVID-19.   \cite{Dagan2020score}. NHS England defined risk factors to predict death from COVID-19 \cite{williamson2020opensafely}. We suggest using a risk score to prioritize COVID-19 testing as part of a ranking system.

\subsection{Monitoring as a Sampling Problem }
Monitoring database activity is based on collecting the data into log files feeding these files to the anomaly detection system. The anomaly detection system uses the data collected to detect anomalies. Database activity is characterized by a vast amount of traffic flow. Since data storage is expensive and limited, the high volumes make it impossible to process and log all of the traffic flows resulting with examining only a sample of the data. as a common solution, such systems use manually defined policies made of rules and activity groups to decide which transactions to save.  

The fact that the policy is manually defined by the company's security expert leads to monitoring that is skewed to the expert's risk scoring method. If exists a risk that the expert is unaware of, this risk might be detected late during the attack, when the cost is high and it is harder to control the damage done . 
Taking a risk driven monitoring approach restricts the security expert's ability to discover concept drifts,such as changes in users' roles, and may cause the “filter bubble” phenomenon, in which the expert is examining only a subspace of events that are over-fitting to the known risks, thereby losing the ability to explore beyond what they already know \cite{nguyen2014exploring,chen2019serendipity}. 

More over because the process of defining the monitoring policy is hard and time consuming, policies have a static nature : once defined, they are rarely changed.

Grushka \textit{et al.} \cite{grushka2020using} suggested taking a dynamic approach by adding diversity into the monitoring policy to add the ability of exploration for new risks characteristics and demographics. They showed the need to balance exploration of activities and users of low risk with exploitation  of high risk scored users and activities. If we put too much emphasis on exploration, we are scouring the space well, but at the cost of scattering and possibly late identification or even missing important events of high risk nature.

\section{Bandit Framework for Epidemic Control}
We suggest a testing framework in which by the end of each predefined time-frame (it could be day or maybe a week) we reexamine our model and adjust it to the new detected demographics and risks in a principled manner, keeping the experts in the loop, and providing them the data needed to discover and explore for new infected demographics and features. We will address a scenario in which the time-frame is defined to be one day. 

This framework is made of daily calibration and  training of the risk scoring model. 
The model is using the risk score to prioritize the potential cases. The majority of the testing budget is invested in the cases with risk score above a thresh-hold, the remainder of the budget is used for collecting new data where there is a lack of data or data that has aged. This exploration motive could be used for exploring  geographic location, demographics,  or exploration of  symptoms that are currently not used in the model. For example, if the doctors are reporting of a symptom that is unknown or not in the model, the experts could set it as an arm that could be pooled using a Thompson sampling. The exploration portion of the tests can also be used as an intervention method. For example, testing all the residents and staff of a nursing home once a case is discovered. 
The experts' knowledge may be used for building the primary risk score model and introducing new possible features to the model or new arms that should be pooled.  

\begin{figure}[ht]
\includegraphics[scale=0.65]{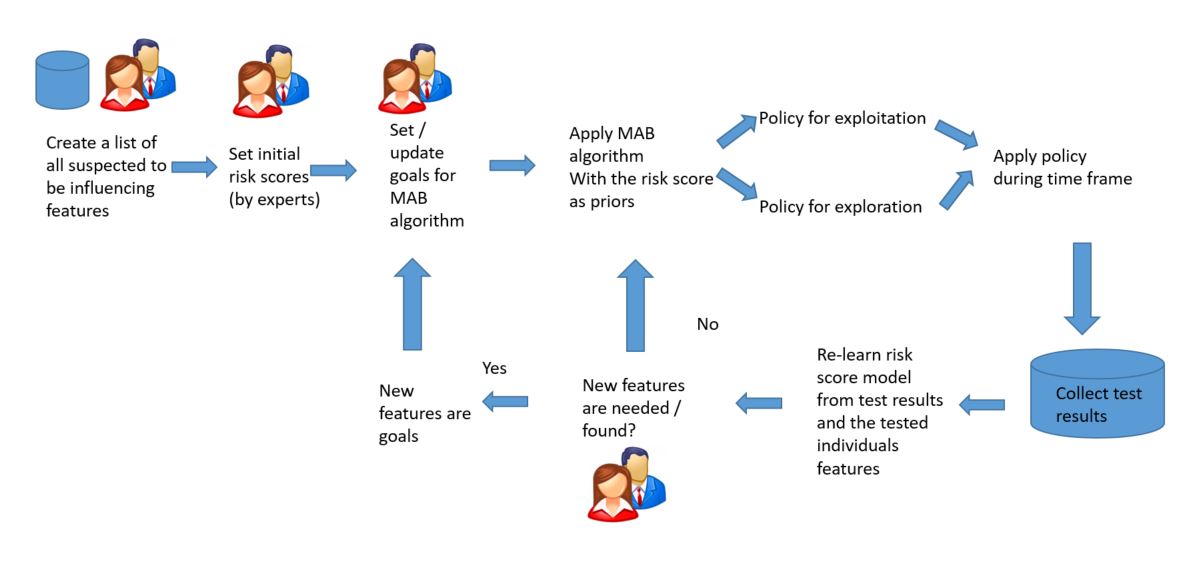}

  \caption{\textbf{MAB Testing Framework workflow} : Each time frame (day/week) the population is tested. A portion of the tests goes to the most likely infected individuals (exploitation) and the rest for exploration (random sampling, or any other expert defined population). The initial risk score can be manually defined based on the epidemiologists knowledge and supported by the data collected by the health organizations. At the end of  each time frame we incorporate the test results and their features, combined with the new results obtained in the field and use this data to  retrain the risk-scoring model (this can be limited to the randomly sampled population to avoid the filter bubble). This framework allows the experts and decision makers to create an explainable and agile policy in a data driven fashion.}
  \label{testing_framework}
  \end{figure}


\section{Methods}
To demonstrate the feasibility of our proposed framework we explore data released by the Israeli Ministry of Health (IMOH) in order to show that effective ranking can be achieved using a simple classifier trained on previous data.

To evaluate the effect of training on data gathered using different testing policies we train another model on data from later weeks (beginning of the second wave of COVID-19 infections) and test both models on a later test set.

\subsection{Data}
To evaluate our proposed thesis we use the Israeli Ministry of Health "tested individuals" COVID-19 data set \cite{govilcovid19}. This data set contains the result of all primary COVID-19 tests by dates collected over 10 weeks from March-11th,2020 until May-10th, 2020 and contains over 300,000 records . Each record represents a test for COVID-19 and is made of the fields: date, partial list of symptoms, including fever, cough, headache, shortness of breath and sore throat, indication for the test, gender and the result of the test (positive\ negative \ other). Individuals with a positive test result will appear only once in the data set, at the date of the first positive result. The indication for the test is a descriptive feature that indicates the reason for taking the test in the first place. This field values are clustered into three categories: (i) contact with confirmed known positive individual (ii) return from abroad and (iii) other. 

This data contains all test preformed including random testing, and testing medical stuff at hospitals. 
our data exploration showed that contact with a positive patient, cough, fever are the strongest features and respectively has the highest correlation with positive test result. 

\subsection{Study Design and Population}
Overall, the data contains 300,456 tested individuals partial symptoms and test results collected between March-11th, 2020 until May-10th, 2020. the data contains null values mainly in the symptoms features, a feature value is set to null if that data wasn't collected. we compared discarding records with null values to referring to null values as if there is no indication for this symptom and decided to take the second approach. 
From the data analysis, we learn that most of the relevant data lie between weeks eleven to sixteen. afterwards, the results show that there is a decrease in the rate of positive results

\begin{figure}[ht]
\centering
    \begin{subfigure}[h]{0.495\textwidth}
        \includegraphics[width=\textwidth]{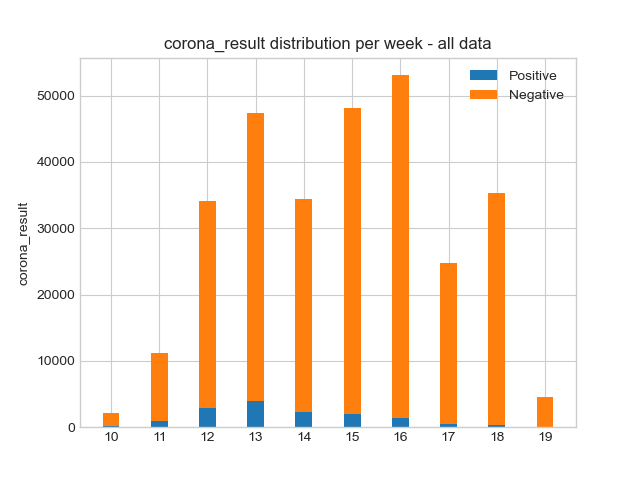}
        \caption{Test results weekly distribution}
        \label{fig:fig1}
    \end{subfigure}
        \begin{subfigure}[h]{0.495\textwidth}
        \includegraphics[width=\textwidth]{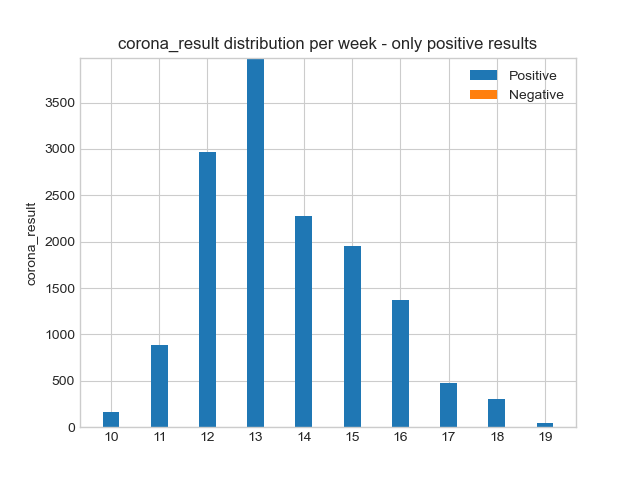}
        \caption{Positive results weekly distribution}
        \label{fig:fig2}
    \end{subfigure}

\caption{Weekly test results distribution}
\label{results distribution}
\end{figure}

\subsection{Features Correlation to Positive Test Result}
To gain insight into the features that contribute most to the predicted probability of being diagnosed with COVID-19, we analyzed feature contribution using Pearson's correlation. 
\begin{table}[ht]  
\centering
\begin{tabular}{||l c||} 
 \hline
 Feature & Median correlation to positive result \\ [0.5ex] 
 \hline\hline
Contact with confirmed & 0.587\\
head ache & 0.292\\
fever & 0.129\\
cough & 0.124\\
sore throat & 0.094\\
shortness of breath & 0.084\\
Abroad & -0.002\\
female (gender) & -0.015\\
Other & -0.433\\[1ex]

\hline
\end{tabular}
\caption{Features correlation to Positive test result}
\label{correlation_table}
\end{table}
Table \ref{correlation_table} shows that the features with the highest positive correlation are: contact with confirmed individual, head ache fever. 

\subsection{Ranking Candidates}
Due to the restricted amount of available tests not all individuals can be tested each day. therefore governments and decision makers needs to set up a policy to define Who should we test first. 
we propose taking the conventional approach from the Cyber security domain of ranking the testing candidates and testing the top X ranked individuals. 
we compare the following policies for ranking candidates for testing in terms of recall and precision at preforming  1000, 2000,3000, 4000 or 5000 tests per week (up to about a tenth of the actual tests performed): 
\begin{itemize}
    \item \textbf{Rule Based} our results show that contact with a positive patient, cough, fever are the strongest features therefore we define a rule based baseline: score them with weights of 2,1,1 respectively
    \item \textbf{Linear support vector machine (SVM)} ranking based on the probability of the linear svm classifier to classify the individual as positive. 
    \item \textbf{Polynomial SVM}  ranking based on the probability of the Polynomial svm classifier to classify the individual as positive. Polynomial of 2 degrees– we assume that it could learn from the relation between two features without over-fitting
\end{itemize}

\subsection{Calculating Confidence Interval}
Confidence interval was calculated using bootstrapping: the population was re-sampled for each week and the recall was calculated for 10 such populations. The confidence interval was calculated using the Student's t distribution as the critical value.

\subsection{Balancing Exploitation (Recall) / Exploration}
We assume that decision makers will not adopt a sampling policy with low recall, as this leaves many patients not diagnosed. In order to find a good balance between exploration and exploitation we examine the recall gained by using 30/40/50/60/70 percent out of the weekly capacity as the sample sizes.  

\subsection{Effect of Retraining the Model on a Different Dataset}
The testing policies affect the resulting data used to train the ranking classifier. We quantified this effect by training two Polynomial-SVM models on different time frames of the data (weeks 10-12 vs weeks 21-23) and evaluated both on a later time frame (weeks 24-26). 

\section{Results}
The data exploration clearly shows that the pandemic was active during weeks 11 to 16

\subsection{Predictive Phenotypes}
We analysed the data to gain better understanding of the strongest predictive phenotypes. The results of the Pearson correlation reflects the testing policy. From the results we can see that the best predictors for positive test results are : contact with infected individual, fever and cough.

\begin{figure}[t]
\includegraphics[scale=0.5]{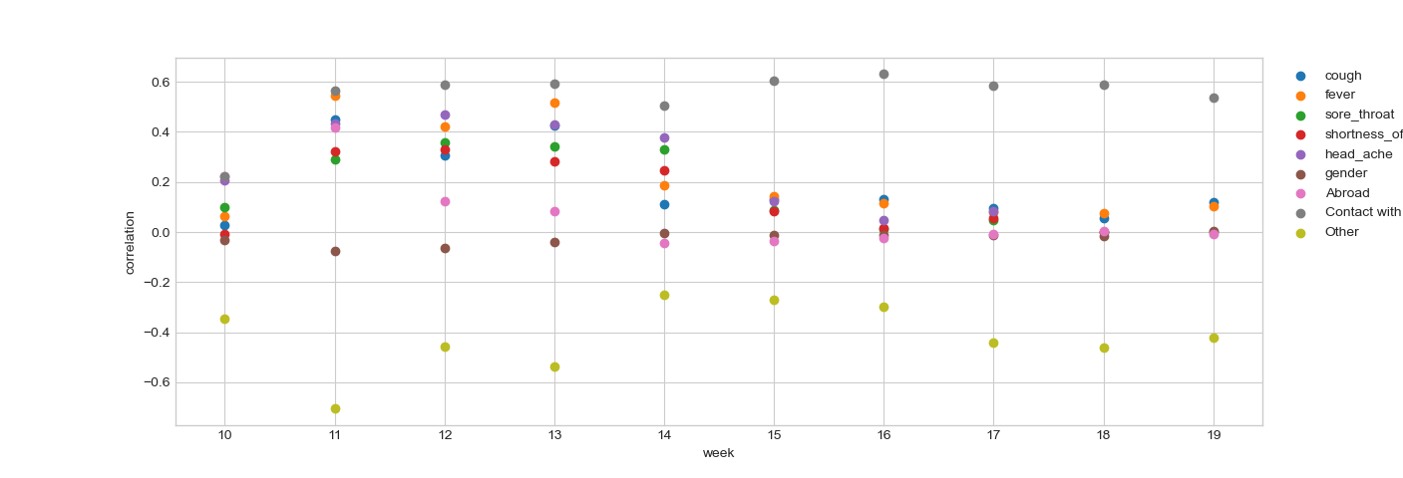}

  \caption{Pearson correlation for each phenotype positive test result by test week}
  \label{features_correlation}
  \end{figure}

\subsection{Ranking Patients Using ML}


\begin{table}[htbp]
\centering
\begin{tabular}{||c | c | c | c | c | c | c||} 
 \hline
 week & recall @1000 & recall @2000 & recall @3000 & recall @4000 & recall @5000 & number of tests \\ [0.5ex] 
 \hline\hline
 13 & 0.231 & 0.429 & 0.623 & 0.667 & 0.732 & 47310  \\ 
 14 & 0.267 & 0.502 & 0.618 & 0.689 & 0.696 & 34340 \\
 15 & 0.369 & 0.565 & 0.585 & 0.596 & 0.609 & 48087 \\
 16 & 0.509 & 0.590 & 0.608 & 0.613 & 0.622 & 53065 \\[1ex] 
 \hline
\end{tabular}
\caption{weekly recall polinamial svm ranking model}
\label{table:Recall_polynomial_svm}
\end{table}



\begin{table}[ht]
\centering
\begin{tabular}{||l | c | c | c | c | c||} 
 \hline
 model & recall @1000 & recall @2000 & recall @3000 & recall @4000 & recall @5000  \\ [0.5ex] 
 \hline\hline
svm-poly kernel & \textbf{0.344}   & \textbf{ 0.521}  & \textbf{0.609} & \textbf{0.657} & 0.666 \\
svm - confidence interval& \texttt{(0.31, 0.379)} &  \texttt{(0.503, 0.542)} & \texttt{(0.602, 0.614)} & \texttt{(0.639, 0.675)} & \texttt{(0.647, 0.683)}\\
linear SVM & 0.317 & 0.517 & 0.606 & 0.619 & 0.623\\
rule-based & 0.269 & 0.412 & 0.522 & 0.629 & \textbf{0.692}\\[1ex]
\hline
\end{tabular}
\caption{Average recall by model}
\label{table:AVG_Recall}
\end{table}

\begin{table}[ht]
\centering
\begin{tabular}{||l c c c c c||} 
 \hline
 model & F @1000 & F  @2000 & F  @3000 & F  @4000 & F  @5000  \\ [0.5ex] 
 \hline\hline
svm-poly kernel & 0.455 & 0.536 & 0.522 & 0.483 & 0.425\\
linear SVM & 0.415 & 0.530 & 0.520 & 0.453 & 0.395\\[1ex]
\hline
\end{tabular}
\caption{Average F1 score by model}
\label{table:AVG_F1}
\end{table}

We compare the models in terms of precision, recall and F1 score. 
Table \ref{table:Recall_polynomial_svm} presents the average weekly recalls based on the the ranking model trained during weeks 10 to 12. We find  \ref{table:Recall_polynomial_svm} that using only $\sim$
10\%  of the testing budget we have a recall of 66.6\% (CI: 64.7\%-68.3\%) of all positive result of each week. The recall of the polynomial SVM at 70\% of the testing capacity was 92.1\%  (CI: 91.1\%-93.2\%).
In terms of F1-score the polynomial SVM out-preforms the linear and rule based models (see Table \ref{table:AVG_F1}). 

\subsection{Effect of Retraining the Model on a Different Dataset}
\begin{figure}[h!]
\includegraphics[scale=0.7]{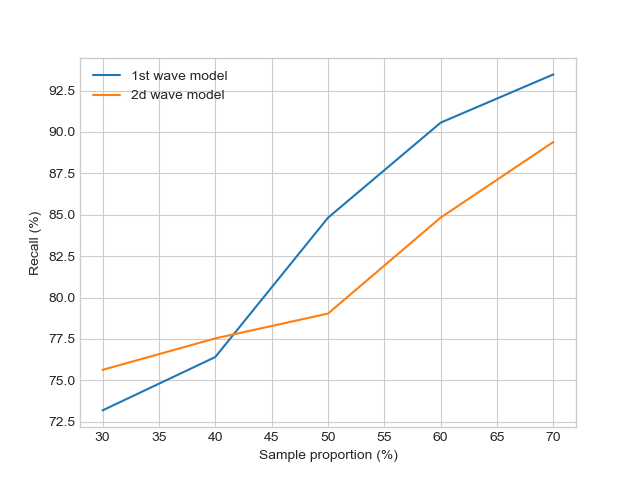}

  \caption{Recall of models trained on different subsets of the data when ranking the patients in weeks 24-26}
  \label{recalls_models}
  \end{figure}

We trained a second polynomial SVM models on weeks 21-23, the period of time where infections started rising again in the dataset. We evaluated both models, 1st wave model and 2nd wave model, on the data of weeks 24-26. We find that the 2nd wave model behaves differently. Recalls at lower capacities were better with the second model and with higher capacities the first model achieved better results (see Figure \ref{recalls_models}).

\section{Discussion}
In this study we suggest a principled data driven MAB framework for maintaining tests policy for the COVID-19 pandemic. Our framework is based on two components, ranking component for maximizing detection and exploration component for maximizing knowledge of the disease in the population. This approach allows the experts to make the high level decision on the policy while preserving the data discovery for re-evaluation. Exploration can be random or based on more complicated sampling schemes such as based on geographic features maximizing information for multiple objectives (based Thompson sampling) to overcome the filter bubble phenomena. Testing policies used by governments can be seen as private cases of this framework: testing only hospitalized patients is choosing maximize detection (100\% exploitation, 0\% exploration) while the policy in Iceland which used two exploration strategies is on the other hand (allocating 10-50\% of the capacity for exploration and basing the exploitation on a manual ranking model). In both of these cases using a data driven ranking model would improve recall (percent of positive tests).

We showed that ranking is feasible using a simple predictive model for predicting positive test results. This corroborates similar finding by \cite{shoer2020should}. Using the data released by the Israel Health Ministry we found ranking could have recall of 66.6\% (CI: 64.7\%-68.3\%) on average using only 10\% of the tests capacity and 92.1\% (CI: 91.1\%-93.2\%) using 70\% of the tests capacity.  A simple rule based model based on the correlations achieved similar accuracy, suggesting that drawing an easy to explain model is possible (at the cost of a very little efficiency). Basing a policy solely on a ranking model would be very efficient capacity wise but would lock the decision makers in a filter bubble. This approach is based on solutions to similar problems where test capacity is limited in the realm of security \cite{grushka2020using}. Saving a major part of the testing capacity for exploration allows both informed decision making and updating the ranking model daily or weekly to maximize the efficiency.

Training a second model using the same architecture on data collected in weeks 21-23 and testing both models on weeks 24-26 we found significant difference in model recall. This is not a surprising results as it is expected that different testing policies would bias the distribution of individuals identified. However, these results show that dedicating a significant portion of the testing capacity to random sampling would provide a clearer picture to decision makers. Furthermore, training such a model once is not good enough, and a framework to support retraining the model and choosing the best ranking model based on data is necessary. In our example, if dedicating 30\% of the testing capacity to exploration it is advisable to use the first wave model, if exploration is raised to 60\% the second model would achieve better recall (see Figure \ref{recalls_models}).

Future work should focus on simulating data using an Agent Based Modeling (ABM) approach and testing different exploration policies under different conditions such as few cases in the population vs. many or geographic sampling vs completely random exploration. We hope to see this framework replace ad-hoc approaches to exploration and testing policy.

\bibliographystyle{splncs04}

\bibliography{covidbib}

\end{document}